\documentclass{article}

\usepackage{PRIMEarxiv}

\usepackage[utf8]{inputenc} 
\usepackage[T1]{fontenc}    
\usepackage{hyperref}       
\usepackage{url}            
\usepackage{xcolor}
\usepackage{fontawesome5}
\usepackage{booktabs}       
\usepackage{amsfonts}       
\usepackage{nicefrac}       
\usepackage{microtype}      
\usepackage{lipsum}
\usepackage{fancyhdr}       
\usepackage{graphicx}       
\usepackage{subcaption}
\usepackage{algorithm}
\usepackage{algpseudocode}
\usepackage{hyperref}
\graphicspath{{media/}}     

\title{AugmenTory: A Fast and Flexible Polygon Augmentation Library
\thanks{\textit{When using this library in your research, we will be happy if you cite us! (or at least bring us some self-made pizza or burgers) \\ \underline{Citation:}}
\textbf{Ghahremani, T., \& Hoseyni, M., \& Ahmadi, M. J., \& Mehrabi, P., \& Nikoofard, A. (2024). AugmenTory: A Fast and Flexible Polygon Augmentation Tool. arXiv preprint arXiv:5581782.}} 
}

\author{
    \textbf{Tanaz Ghahremani} \textsuperscript{}\\
    SmartoryLabs \\
    \texttt{\href{mailto:tanaz@smartory.co}{tanaz@smartory.co}} \\
    \and
    \textbf{Mohammad Hoseyni}
    \textsuperscript{}\\
    SmartoryLabs \\
    \texttt{\href{mailto:mohammad@smartory.co}{mohammad@smartory.co}} \\
    \and
    \textbf{Mohammad Javad Ahmadi}\\
    SmartoryLabs \\
    \texttt{\href{mailto:mjahmadi@smartory.co}{mjahmadi@smartory.co}} \\
    \and
    \textbf{Pouria Mehrabi}\\
    SmartoryLabs \\
    \texttt{\href{mailto:pouria@smartory.co}{pouria@smartory.co}} \\
    \and
    \textbf{Amirhossein Nikoofard}\\
    SmartoryLabs \\
    \texttt{\href{mailto:nikoofard@smartory.co}{nikoofard@smartory.co}} \\
}
\begin{document}
\maketitle

\begin{abstract}
Data augmentation is a key technique for addressing the challenge of limited datasets, which have become a major component in the training procedures of image processing. Techniques such as geometric transformations and color space adjustments have been thoroughly tested for their ability to artificially expand training datasets and generate semi-realistic data for training purposes.

Polygons play a crucial role in instance segmentation and have seen a surge in use across advanced models, such as YOLOv8. Despite their growing popularity, the lack of specialized libraries hampers the polygon-augmentation process. This paper introduces a novel solution to this challenge, embodied in the newly developed \textit{AugmenTory} library. Notably, \textit{AugmenTory} offers reduced computational demands in both time and space compared to existing methods. Additionally, the library includes a postprocessing thresholding feature. The \textit{AugmenTory} package is publicly available on GitHub, where interested users can access the source code: 
\href{https://github.com/Smartory/AugmenTory}{github.com/Smartory/AugmenTory}
\end{abstract}


\section{Introduction}
Artificial intelligence (AI) is rapidly advancing in various domains, including computer vision. The effectiveness of AI largely depends on the quality and quantity of data, which are crucial for enhancing performance in this field \cite{yang2022image}. 
Data augmentation is a valuable strategy to mitigate issues related to limited datasets. Not only does it address data scarcity, but also it boosts the model's robustness and generalizability. Specifically, rule-based data augmentation, which includes techniques like rotation, flipping, and cropping, systematically modifies existing data based on predefined rules \cite{choi2024softeda}.

Segmentation is a fundamental task in AI, focusing on isolating an object's area within an image to pinpoint its exact location. This capability is vital for applications across various industries. Ground truth in segmentation involves two primary methodologies. The first method, known as pixel-wise classification, is essential in fields such as remote sensing and image processing. This technique assigns each pixel a semantic class within a multi-channel image format, where each channel represents a different class. Thus, for each identified object, there is a dedicated channel in the ground truth image \cite{guo2018pixel}. This method provides a detailed and precise representation of the image and is thus invaluable for tasks that require high accuracy. Such detailed segmentation ground truth is commonly referred to as a 'Mask'.

The other method is to use a set of coordinates for determining an object's shape; this method has shown significant efficacy in segmentation tasks within neural network-based image processing. It improves models' accuracy in identifying individual instances of a class \cite{hafiz2020survey}. In contrast to mask annotation, which uses image masks, this technique captures a coordinate array for each object, requiring less storage and making it usable on systems with limited disk and RAM capacity. It is more efficient and less resource-demanding than the detailed but resource-intensive pixel-wise segmentation methods, providing an ideal balance of detail and computation efficiency. This makes it ideal for applications like autonomous vehicle navigation, medical imaging, and aerial crop monitoring, which require precise object boundaries without pixel-level accuracy. Furthermore, its adaptability makes it ideal for marking irregularly shaped objects, as it can conform more easily to complex outlines than bounding boxes or other simple geometric shapes. In the field, this method is commonly known as 'Polygon' annotation.

There are several well-established libraries designed to augment mask annotations, such as \textit{Albumentations}. However, augmenting polygon annotations remains a significant challenge for researchers in the AI field. To address this gap and inspired by \textit{Albumentations}, we developed a specialized library, that focuses on geometric augmentations. This new library offers efficient solutions in terms of time and space complexity compared to existing methods. Additionally, it includes a thresholding post-process that manages the overlap of objects after transformation. To our knowledge, this is the first study that applies augmentation algorithms specifically for polygon annotations.
\begin{figure}[htbp]
    \centering
    
    \begin{subfigure}[t]{0.45\textwidth} 
        \centering
        \includegraphics[width=\textwidth, height=5cm, keepaspectratio]{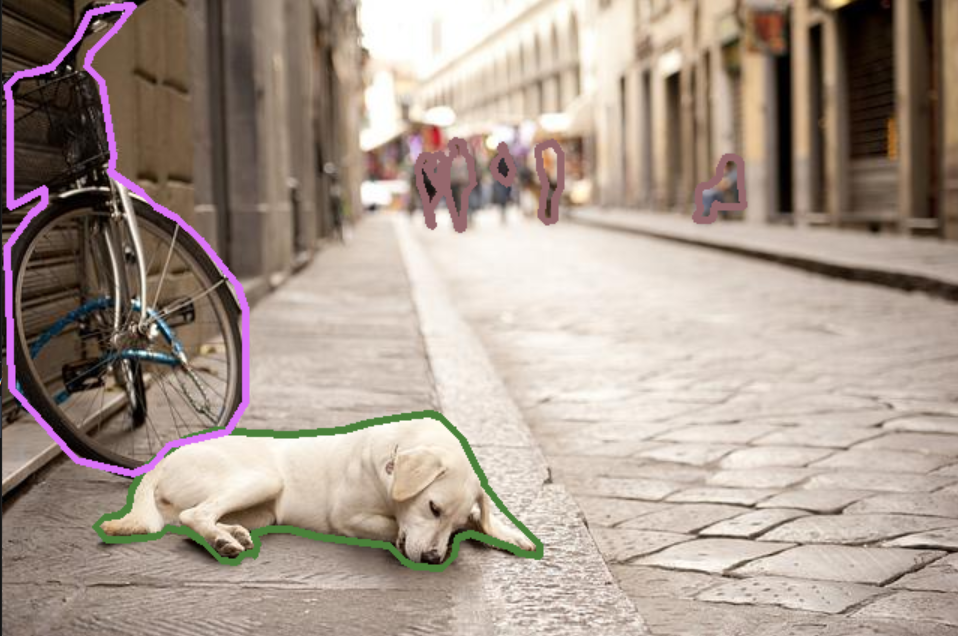} 
        \caption{Original image}
        \label{fig:OriginalCOCO}
    \end{subfigure}
    \quad 
    \begin{subfigure}[t]{0.45\textwidth} 
        \centering
        \includegraphics[width=\textwidth, height=5cm, keepaspectratio]{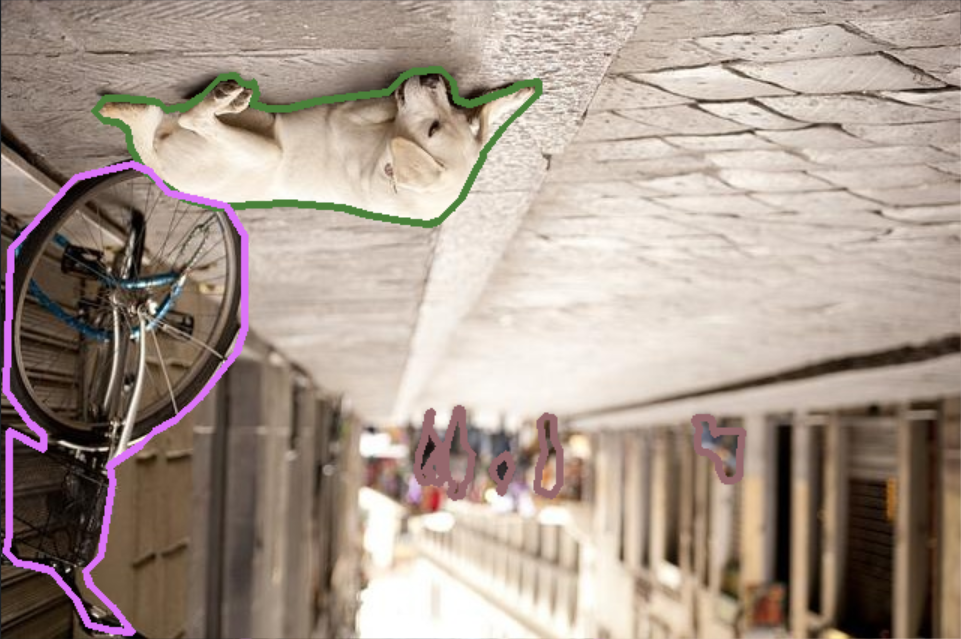} 
        \caption{Vertical Flipped}
        \label{fig:VerticalFlipCOCO}
    \end{subfigure}
    
    \vspace{1em} 
    
    \begin{subfigure}[t]{0.45\textwidth} 
        \centering
        \includegraphics[width=\textwidth, height=5cm, keepaspectratio]{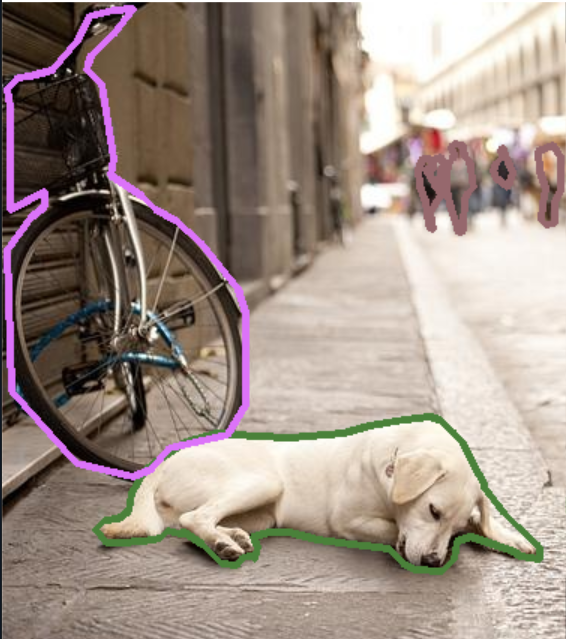} 
        \caption{Cropped}
        \label{fig:CropCOCO}
    \end{subfigure}
    \quad 
    \begin{subfigure}[t]{0.45\textwidth} 
        \centering
        \includegraphics[width=\textwidth, height=5cm, keepaspectratio]{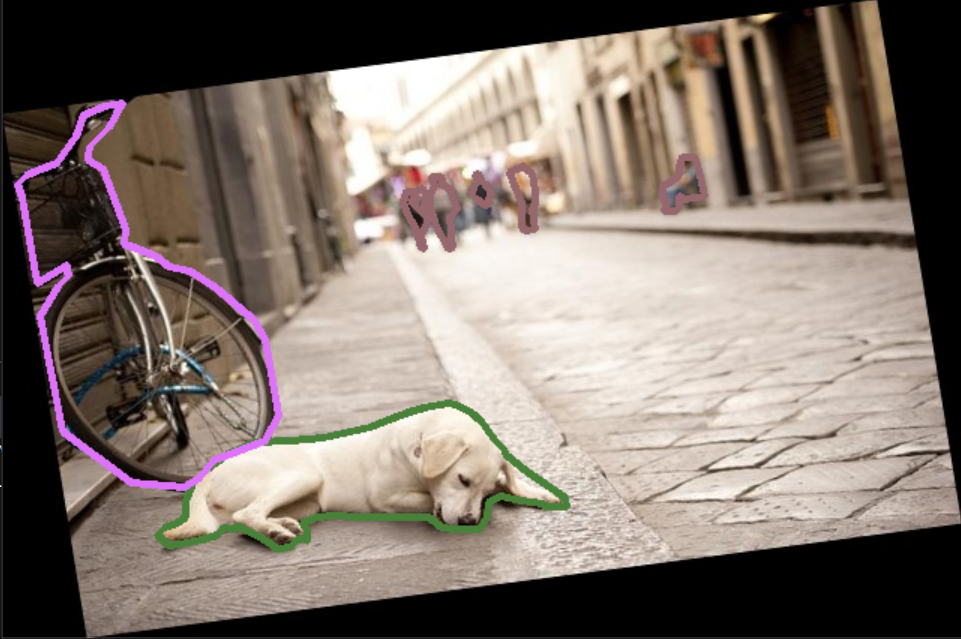} 
        \caption{Rotated}
        \label{fig:RotateCOCO}
    \end{subfigure}
    
    \caption{The experiment on COCO dataset with the threshold of 0}
    \label{fig:COCO}
\end{figure}
\section{Methodology}
The \textit{AugmenTory} introduces an innovative approach to polygon augmentation, in which each vertex coordinate pair is treated as a distinct keypoint. In this method, every coordinate pair within the polygon is allocated a unique identifier, referred to as the keypoint's name. Subsequent to this identification, keypoints undergo a transformation process utilizing an \textit{Albumentations}' transform. The altered keypoints are subsequently reconstructed into their respective polygons using the assigned identifiers. However, certain transformations, such as rotations or crops, may displace keypoints beyond the image boundaries. To address this issue, the Intersection over Union (IoU) between the polygon's pre- and post-transformation states is computed. If the IoU falls below a predefined threshold, the transformed instance is disregarded. This methodology ensures the integrity and applicability of the polygon within the transformed image context.

\begin{algorithm}
\caption{Polygon Transformation with Albumentations}
\label{alg: SudoCode}

\begin{algorithmic}[1]
\State \textbf{Input:} Original polygons, image, Albumentations transform
\State \textbf{Output:} Transformed images and polygons

\Procedure{YOLO2KeyPoint}{ image, labels}
    \State Read labels file
    \State Compute the area of each polygon
    \State Encode each polygon into a set of keypoints with unique names 
    (\verb|ID_CLASS_AREA|)
    \State \Return keypoints, class labels
\EndProcedure

\Procedure{keyPoint2YOLO}{transformed\_key\_points, transformed\_class\_labels, overlap}
    \State Decode the transformed keypoints to recreate polygons
    \For{each polygon}
        \State Calculate the area of the polygon
        \State Check if the polygon is significantly outside the image boundaries based on the \verb|overlap|
        \If{outside area is above a threshold}
            \State Dismiss the polygon
        \Else
            \State Keep the polygon for further processing
        \EndIf
    \EndFor
    \State \Return processed polygons
\EndProcedure

\Procedure{do\_aug\_poly}{transformation, image\_addr, image\_labels, save\_path}

    \Call{YOLO2KeyPoint}{image, labels}
    \State Apply Albumentations transform to these keypoints along with the image
    
    \Call{keyPoint2YOLO}{transformed\_key\_points, transformed\_class\_labels, overlap}
    \State Save transformed\_image and transformed\_label
\EndProcedure

\State \textbf{Begin Main Procedure}
\State \textbf{Inputs:} Load original polygons and image
\State \textbf{Transform:} Specify Albumentations transform parameters
\State transformed\_key\_points, transformed\_image $\gets$ \Call{do\_aug\_poly}{transformation, image\_addr, image\_labels, save\_path}
\State \textbf{End Main Procedure}

\end{algorithmic}
\end{algorithm}

The algorithm, detailed in \autoref{alg: SudoCode}, outlines how the \textit{AugmenTory} uses \textit{Albumentations} to transform polygons. The first step, YOLO2KEYPOINT, requires an image and the label file as inputs. The system reads the label file and extracts polygons in YOLO format. It then calculates the area of each polygon. In the final step of this process, each polygon point is assigned a unique ID, class, and area before being converted into a set of keypoints. The YOLO2KEYPOINT procedure eventually returns these keypoints. 

The next process, KEYPOINT2YOLO, accepts transformed keypoints, class labels, and an overlap threshold as inputs. It converts the keypoints back into polygon points using each keypoint's unique identifiers, effectively decoding a set of polygons from the keys. The algorithm calculates each polygon's area and compares it to its original dimensions. Polygons with an area less than the specified threshold are discarded.
The concluding process, DO\_AUG\_POLY, integrates both YOLO2KEYPOINT and KEYPOINT2YOLO with a specific transformation. It processes the images and labels accordingly and stores the transformed results.

\section{Experiment}

The \textit{AugmenTory} is compatible with all \textit{Albumentation}-related transformations that apply to keypoints. To assess its functionality, an experiment was conducted using an image from the COCO benchmark, as illustrated in \autoref{fig:COCO}. The original image selected for augmentation, shown in \autoref{fig:OriginalCOCO}, included various classes such as dog, bicycle, and human, with the human class containing multiple objects. The applied transformations were a vertical flip \autoref{fig:VerticalFlipCOCO}, a crop \autoref{fig:CropCOCO}, and a rotation \autoref{fig:RotateCOCO}, all executed with a 0\% Intersection over Union (IoU) threshold.

\begin{figure}[htbp]
    \centering
    \begin{subfigure}[b]{\textwidth}
        \centering
        \includegraphics[width=7cm]{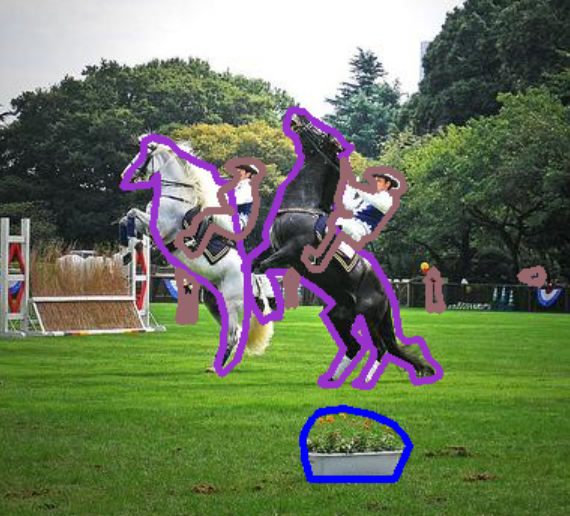}
        \caption{Original Image}
        \label{fig: HorseOriginal}
    \end{subfigure}
    
    \begin{subfigure}[b]{0.33\textwidth}
        \centering
        \includegraphics[width=3.5cm]{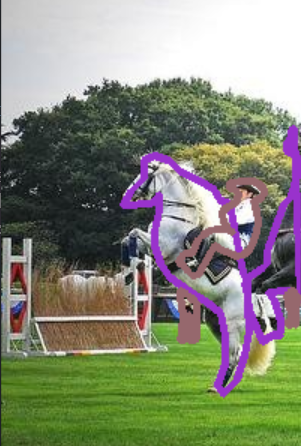}
        \caption{Cropped with 0\% Threshold}
        \label{fig: HorseCrop0}
    \end{subfigure}
    \begin{subfigure}[b]{0.33\textwidth}
        \centering
        \includegraphics[width=3.5cm]{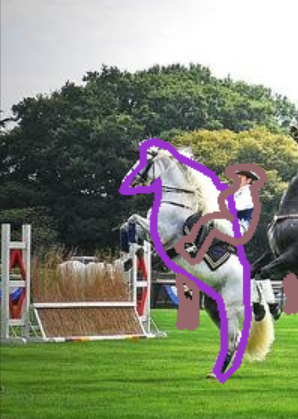}
        \caption{Cropped with 20\% Threshold}
        \label{fig: HorseCrop20}
    \end{subfigure}

    \caption{The experiment of COCO dataset while \autoref{fig: HorseCrop0} has 0\% overlap and \autoref{fig: HorseCrop20} has 20\% overlap for post-process threshold.}
    \label{fig: Horse}
\end{figure}

A secondary experiment was designed to assess the performance of thresholds in the post-process transformation, as depicted in \autoref{fig: Horse}, which is another image from the COCO-128-seg benchmark dataset. \autoref{fig: HorseOriginal} displays three distinct classes of horse, human, and flower, with respective instance counts of two, six, and one.  Utilizing the AugmenTory library, the original image underwent two cropping processes with identical coordinates but varying thresholds. The image in \autoref{fig: HorseCrop0} was generated with a 0\% threshold, retaining the polygon of the black horse as an independent object; as a result, this object remains indistinguishable to deep learning models. Conversely, the image in \autoref{fig: HorseCrop20} was produced using a 20\% threshold, which excludes objects with less than 20\% IoU pre-transformation; hence, the polygon of the black horse was omitted. This approach streamlines the training process by focusing on more distinct and recognizable features, thereby enhancing the model’s learning efficacy and its ability to generalize across well-defined objects.

The methodology introduced in this study is carefully designed to enhance the efficiency of the transformation algorithm, significantly optimizing both time and space complexities. \autoref{table1: comparision} provides a comparative analysis between the proposed algorithm and the conventional approach for transferring polygon annotations. The first row indicates the time and space required for executing a vertical flip transformation on the COCO-128-seg dataset, a recognized benchmark. Remarkably, the algorithm utilized in this study requires only 1.2\% of the space demanded by traditional methods, with a time saving of 3.907 seconds. Additionally, the second row documents the average time and space consumption for ten identical operations involving the rotation transformation on \autoref{fig:OriginalCOCO}, highlighting the superior speed and resource efficiency of the proposed algorithm.

\begin{table}[htbp]
    \centering
    \begin{tabular}{l|ll|ll}
                         & \multicolumn{2}{c|}{Mask}                             & \multicolumn{2}{c}{Polygon}                          \\ \hline
    Experiment           & \multicolumn{1}{c}{Time} & \multicolumn{1}{c|}{Space} & \multicolumn{1}{c}{Time} & \multicolumn{1}{c}{Space} \\ \hline
    COCO 128-seg Dataset & 9.4910 s                   & 225.835 MB& 5.5840 s                   & 2.802 MB\\
    Ours (\autoref{fig:OriginalCOCO})                  & 0.0820 s 
    & 36.915 MB& 0.0506 s 
    & 1.868 MB\end{tabular}
    \caption{The comparison between time complexity and space complexity of the proposed approach and conventional method.}
    \label{table1: comparision}
\end{table}
\section{Discussion and Conclusion}
The transformation of polygon annotations presents significant challenges in the field of computer vision, due to a scarcity of specialized libraries. Traditionally, this process involves converting each polygon into a corresponding mask, applying the desired transformation with a tool like \textit{Albumentations}, and then reconstructing a polygon that resembles the shape of the transformed mask. This traditional approach, however, is complex and particularly cumbersome when used to augment images for instance segmentation.
In contrast, the recently released \textit{AugmenTory} library offers a more streamlined and efficient solution. It applies \textit{Albumentations'} comprehensive suite of transformations directly to keypoints, allowing these transformations to be used with polygon annotation. This method significantly increases processing speed while reducing data storage requirements. The time and space complexity of \textit{AugmenTory's} transformations is determined by the specific \textit{Albumentations} transform used, emphasizing the library's flexible integration capabilities. \textit{AugmenTory} also includes a thresholding feature that evaluates the overlap of transformed objects, discarding annotations where the overlap falls below a predetermined threshold.
Furthermore, the proposed library, \textit{AugmenTory}, is built with flexibility in mind. It is easily adaptable to work with any other library that supports augmentation transforms, such as \textit{PyTorch} or \textit{TensorFlow}. This adaptability makes \textit{AugmenTory} a versatile tool that can be incorporated into a variety of computer vision workflows, increasing its utility and efficiency.

In conclusion, the new method provided by \textit{AugmenTory} presents a significant improvement over the traditional approach to the transformation of polygon annotation. By applying \textit{Albumentations’} transformations directly to keypoints, it reduces both time and space complexity, making it a more efficient solution. Furthermore, the thresholding feature allows for more precise control over the results of the transformation. As such, \textit{AugmenTory} represents a valuable tool for computer vision engineers working with polygon annotations.

\bibliographystyle{unsrt}  
\bibliography{references}

\end{document}